# AN OVERVIEW OF HINDI SPEECH RECOGNITION

Neema Mishra
M.Tech. (CSE) Project Student
G H Raisoni College of Engg.
Nagpur University, Nagpur, INDIA
neema.mishra@gmail.com

Urmila Shrawankar
CSE Dept.
G H Raisoni College of Engg.
Nagpur University, Nagpur, INDIA
urmila@ieee.org

Dr. V. M Thakare
Professor & Head
PG Dept. of Computer Science
SGB Amravati University, Amravati, INDIA

*Abstract*

*In this age of information technology, information access in a convenient manner has gained importance. Since speech is a primary mode of communication among human beings, it is natural for people to expect to be able to carry out spoken dialogue with computer [1]. Speech recognition system permits ordinary people to speak to the computer to retrieve information. It is desirable to have a human computer dialogue in local language. Hindi being the most widely spoken Language in India is the natural primary human language candidate for human machine interaction. There are five pairs of vowels in Hindi languages; one member is longer than the other one.  This paper describes an overview of speech recognition system. How speech is produced and the properties and characteristics of Hindi Phoneme.*

## 1 Introduction

In India if it could be possible to provide human like interaction with the machine, the common man will be able to get the benefits of information and communication technologies. In this scenario the acceptance and usability of the information technology by the masses will be tremendously increased. Moreover 70% of the Indian population lives in rural areas so it becomes even more important for them to have speech enabled computer application built in their native language. Here we must mention that in the past time decades, the research has been done on continuous, large vocabulary speech processing systems for English and other European languages; Indian languages as Hindi and others were not being emphasized [2]. India is passing through the phase of computer revolution. Therefore it is time that speech recognition technology must be developed for Indian languages.

Speech recognition refers to the ability to listen (input in audio format) spoken words and identify various sounds present in it, and recognize them as words of some known language. Speech recognition in computer domain involves various steps with issues attached with them. The steps required to make computers perform speech recognition are: Voice recording, word boundary detection, feature extraction, and recognition with the help of knowledge models.

Word boundary detection is the process of identifying the start and the end of a spoken word in the given sound signal. While analyzing the sound signal, at times it becomes difficult to identify the word boundary. This can be attributed to various accents people have, like the duration of the pause they give between words while speaking. Feature Extraction refers to the process of conversion of sound signal to a form suitable for the following stages to use. Feature extraction may include extracting parameters such as amplitude of the signal, energy of frequencies, etc. Recognition involves mapping the given input (in form of various features) to one of the known sounds. This may involve use of various knowledge models for precise identification and ambiguity removal. Knowledge models refers to models such as phone acoustic model, language models, etc. which help the recognition system. To generate the knowledge model one needs to train the system. During the training period one needs to show the system a set of inputs and what outputs they should map to. Through this paper, we describe the development of speech recognition system for Hindi Language.

This paper is organized as follows. In the next section production of speech and signal processing has covered. In section 3, the basic concept of speech recognition is described. Section 4 gives the observations of prosodic characteristics of the Hindi Vowels and consonants. Section 5 describes the conclusion.

### 1.1  Existing System

Although some promising solutions are available for speech synthesis and recognition, most of them are tuned to English.

The acoustic and language model for these systems are for English language. Most of them require a lot of configuration before they can be used. There are also projects which have tried to adapt it to Hindi or other Indian languages. Explains how a acoustic model can be generated using a existing acoustic model for English [3]. SIP [4] and Sphinx [5] are two of the known *Speech Recognition* software in open source. A comparison of public domain software tools for speech recognition [6].Some commercial software like IBM's Via Voice is also available.

## 2  How Speech is produced?

Human speech is produced by vocal organs presented in Figure. **1** the main energy source is the lungs with the diaphragm. The vocal tract opening of the vocal cords, or glottis, and ends at the lips. The vocal tract consists of the pharynx (the connection from the esophagus to the mouth) and the mouth, or the oral cavity. The cross-sectional area of the vocal tract, determined by the position of the tongue,lips,jaw and velum varies from zero to about 20 cm2.The nasal tract begins at the velum and ends at the nostrils. When the velum is lowered, the nasal tract is acoustically coupled to the vocal tract to produce the nasal sounds of speech.

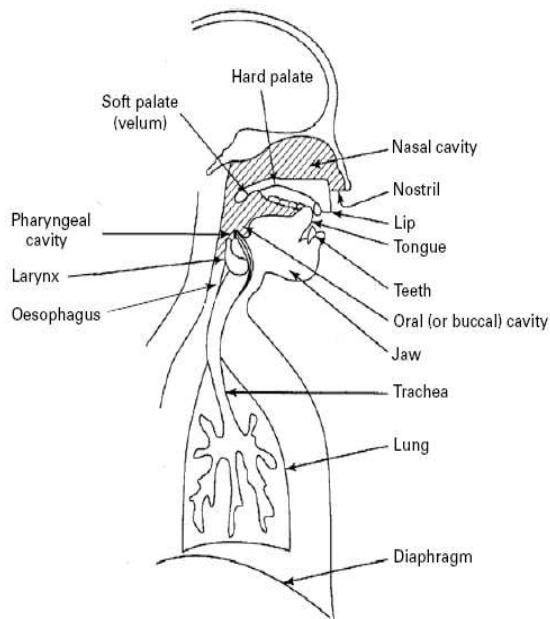

Figure 1.Diagram of the human speech production mechanism

Air enters the lungs via the normal breathing mechanism. As air is expelled front the lungs through the trachea, the tensed vocal cords within the larynx are caused to vibrate by the air flow. The air flow is chopped into quasi-periodic pulses which are then modulated in frequency in passing through the pharynx, the mouth cavity, and possibly the nasal cavity. Depending on the position of the various articulators (i.e., jaw, tongue, velum, lips, mouth), different sounds are produced [7].

## 3  Architecture of Speech Recognition System

Speech recognition is a special case of pattern recognition. Figure 2 shows the processing Stages involved in speech recognition. There are two phases, training and testing. The process of extraction features relevant for classification is common in both phases. During the training phase, the parameters of the classification model are estimated using a large number of class exemplars. During the testing phase, the feature of test pattern is matched with the trained model of each and every class. The test pattern is declared to belong to that class whose model matches the test pattern best.

The goal of speech recognition is to generate the optimal word sequence subject to linguistic constraints. The sentence is composed of linguistic units such as words, syllables, phonemes. In speech recognition, a sentenced model is assumed to be a sequence of models of smaller units. The acoustic evidence provided by the acoustic models of such units is combined with the rules of constructing valid and meaningful sentences in the language to hypothesis the sentences.Therfore, in case of speech recognition, the pattern matching system can be viewed as taking place in two domains: acoustic and symbolic. In the acoustic domain, a featured vector corresponding to a small segment to test speech is matched with the acoustic model of each and every class. The segment is assigned the label of the class with the highest matching score. This process of label assignment is repeated for every feature vector in the feature vector sequence computed from the test data. The resultant sequence of labels is processed in conjunction with the language model to yield the recognized sentenced [8].

The basics structure of speech recognition system is shown in figure 2.

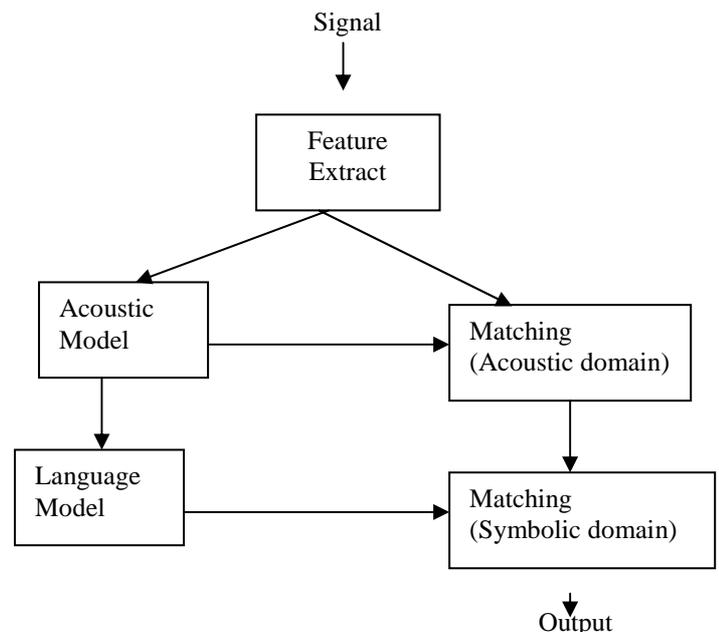

Figure 2. The Block Diagram of a Speech Recognition System

**3.1 Speech Signal**

Speech propagates as a longitudinal wave in a medium, such as air or water. The speed of propagation depends on the density of the medium. It is common to plot the *amplitude* of air pressure variation corresponding to a speech signal as a function of time; this kind of plot is known as a speech pressure waveform or just a speech waveform. Speech signal is an analog signal at the recording time, which varies with time. To process the signal by digital means, it is necessary to sample the continuous-time signal into a discrete-time signal, and then convert the discrete-time continuous valued signal into a discrete-time, discrete-valued (digital) signal. The properties of a signal change relatively slow with time, so that we can divide the speech into a sequence of uncorrelated segments, or frames, and process the sequence as if each frame has fix properties. Under this assumption, we can extract the features of each frame based on the sample inside the frame only. And usually, the feature vector will replace the original signal in the further processing, which means the speech signal is converted from a time varying representing events in the probability space is called Signal Modeling [9] [10].

### 3.2 Feature Extraction

The common step in feature extraction is frequency or spectral analysis. The signal processing techniques aim to extract features that are related to identify the characteristics [11].The goal of feature extraction is to find a set of properties of an utterance that have acoustic correlations in the speech signal i.e. parameters that can some how estimated through processing of signal waveform. Such parameters are termed as features [12].

Several different feature extraction algorithm exist, namely
- Linear Predictive Cepstral Coefficients(LPCC)
- Perceptual Linear Prediction(PLP) Cepstra
- Mel-Frequency Cepstral Coefficients (MFCC)

LPCC computes Spectral envelop before converting it into Cepstral coefficient. The LPCC are LP-derived cepstral coefficient. The PLP integrates critical bands, equal loudness pre emphasis and intensity to compressed loudness. The PLP is based on the Nonlinear Bark scale. The PLP is designed to speech recognition with removing of speaker dependant characteristics. MFCC are extensively in ASR.MFCC is based on signal decomposition with the help of a filter bank, which uses the Mel scale. The MFCC results on Discrete Cosine Transform (DCT) of a real logarithm of a short-time energy expressed on the Mel frequency scale. Most research work on 12MFCC.The Cepstral coefficients are set of feature reported to be robust in some different pattern recognition tasks concerning with human voice. Human voice is very well adapted to the ear sensitivity. In speech recognition, tasks 12 coefficients are retained which represent the slow variation of the spectrum of the signal, which characterizes the vocal tract shape of the uttered words [13]. The Mel-frequency cepstrum coefficient technique is often used to create the fingerprint of the sound files. The mfcc are based on the known variation of the human ear's critical bandwidth frequencies with filter spaced linearly at low frequencies and logarithmically at high frequencies used to capture important characteristics of speech. Studies have shown that human perception of the frequencies contents of sound for speech signal does not follow a linear scale. Thus, for each store with an actual frequency, measures in Hz; a subjective pitch is measured on a scale called the Mel-scale. The Mel frequency scale is below 1000 Hz and a logarithmic spacing above 100Hz.As reference point the pitch of 1 KHz, tone, 40db above the perceptual hearing threshold is defined as 1000 Mels.The following formula is used to compute the Mels for particular frequency:

**Mel (f) = 2595*$\log_{10}$(1+f/700)**

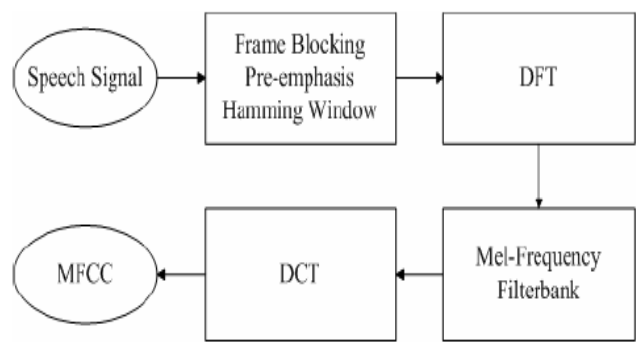

**Figure 3: Block Diagram of the MFCC Processes**

In frame, blocking the speech waveform is cropped to remove silence or acoustical interference that may be present in the beginning or end of the sound file. The windowing block minimizes the discontinuities of the signal by tapering the beginning and end of each frame to zero. The Fast Fourier Transform (FFT) block converts each frame from the time domain to the frequency domain. In the Mel Scale Filter block, the signal is filtered using band-pass filter whose bandwidths and spacing are roughly equal to critical bands and whose range of center frequencies covers frequencies most important for speech perception (300-5000Hz).In Inverse Discrete Fourier Transform (IDFT) block, Cepstrum is generated. Where Cepstrum is the spectrum of the log of the spectrum. MFCC feature is considered for speaker –independent speech recognition and for the speaker recognition task as well [14].The given table describes the comparison of LPCC and MFCC.

| Categorization | MFCC | LPCC |
|---|---|---|
| Low Bandwidth | Higher result | Less Effective |
| Noisy | Effective | Less Effective |
| Vocal Tract | Yes | No |
| Human Ear | Good | Bad |

**Table 1: Comparison between MFCC and LPCC**

### 3.3 Acoustic Modeling

Feature that are extracted by the Feature Extraction module meet to be compared against a model to identify the sound that was produced as the word that was spoken. This model is called as Acoustic model. In this, the acoustic information and phonetics is established. Speech unit is mapped to its acoustic counterpart using model as a speech signal. Most popular acoustic model is Hidden Markov Model (HMM) .There are two types of acoustic models Word Model and Phone Model. Word model are generally used to small vocabulary systems. In this model the words are modeled as whole. Thus, each words needs to be modeled separately. If we need to add support to recognize a new word, we will have to train the system for the word. In the recognition process, the sound is matched against each of the model to find the best match. This best match is assumed to be the spoken word. In Phone Model, instead of modeling the whole word, we model only parts of the words generally phones. The word itself is modeled as sequence of phone. The heard sound is now matched against the parts and parts are recognized. The recognized parts are put together to for a word. For example the word *Ak* is generated by combination of two phones *A* and *k*.This is useful when we need a large vocabulary system. Adding a new word in the vocabulary is easy as the sounds of phones are already know only the possible sequence of phone for the word with it probability needs to be added to the system.

In this, the acoustic information and phonetics is established. Speech unit is mapped to its acoustic counterpart using model as a speech signal. Most popular acoustic model is Hidden Markov Model (HMM).

### 3.3 Language Modeling

The goal of the language model is to produce accurate value of probability of a word W, Pr (w).A language model contains the structural constraints available in the language to generate the probabilities. Although there are words that have similar sounding phone, humans generally do not find it difficult to recognize the word. This is mainly because they know the context, and also have a fairly good idea about what words or phrases can occur in the context. Providing this context to a speech recognition system is the purpose of language model. The language model specifies what are the valid words in the language and in what sequence they can occur. Generally small vocabulary constrained tasks like phone dialing can be modeled by grammar based approach where as large application like broadcast news transcription require stochastic approach.

### 4. Acoustic-Phonetic Feature of Hindi

The acoustic-phonetic of Hindi differs from the European languages. The Hindi alphabet consists of 10 vowels (including 2 diphthongs), 4 semivowel, 4 fricatives and 25 stop consonants (including 5 nasals). The stop consonants are ordered in a systematic manner in most of the Indian language and this order may suggest ideas for developing a recognition system. Here we described characteristic of Hindi Vowel and Consonants. In Table 2. It has three sections: First section consists of vowels, the second section consists of phonemes whose production involves complete closure of oral tract. The third section consists of semivowels and fricatives, in which speech sound is made, while the mouth from one position to another.

The Hindi alphabet consists of 10 vowels (including 2 diphthongs), 4 semivowel, 4 fricatives and 25 stop consonants (including 5 nasals). The stop consonants are ordered in a systematic manner in most of the Indian language and this order may suggest ideas for developing a recognition system. Here we described characteristic of Hindi Vowel and Consonants. Table 2. Shows the arrangement of vowel, consonants and semivowels.

Some examples of the unaspirated consonants with other vowel-endings are give in Table-1.2.The vowel consists of three types. Front, mid and back. The three short vowels are (A), (E) and (U) are basic. The (a), (i), (u) are long vowels. Combination of two vowel called as diphthongs (ai), (au).

The stop consonants are produced by completely closing the oral cavity and then releasing the built-air pressure. The different consonants have different point of closure in the oral cavity. The innermost point of closure in the mouth is at the glotiis.the other points of closure are where the back, middle and front parts of the tongue press against the appropriate regions of the upper palate.Finally, beyond the teeth there is the closure of the lips. Thus the 5 rows in table 1.represent five different classes of consonants. These correspond to the five places of articulation.

| अ | आ | इ | ई | उ | ऊ | ए | ऐ | ओ | औ |
|---|---|---|---|---|---|---|---|---|---|
| $a$ | $a\!:$ | $i$ | $i\!:$ | $u$ | $u\!:$ | $e$ | $e\!:$ | $o$ | $o\!:$ |
| a | A | i | I | u | U | e | E | o | O |

| क | ख | ग | घ | ङ |
|---|---|---|---|---|
| $k$ | $k^h$ | $g$ | $g^h$ | $\eta$ |
| k | kh | g | gh | gñ |
| च | छ | ज | झ | ञ |
| $t\!\int$ | $t\!\int^h$ | ʤ | ʤ$^h$ | $\eta$ |
| c | ch | j | jh | jñ |
| ट | ठ | ड | ढ | ण |
| ʈ | ʈ$^h$ | ɖ | ɖ$^h$ | ɳ |
| T | Th | D | Dh | N |
| त | थ | द | ध | न |
| $t$ | $t^h$ | $d$ | $d^h$ | $n$ |
| t | th | d | dh | n |
| प | फ | ब | भ | म |
| $p$ | $p^h$ | $b$ | $b^h$ | $m$ |
| p | ph | b | bh | m |

| य | र | ल | व | श | ष | स | ह |
|---|---|---|---|---|---|---|---|
| $j$ | $r$ | $l$ | $\omega$ | $\int$ | $\c{s}$ | $s$ | $h$ |
| y | r | l | w | s~ | S | s | h |

**Table 2.The Hindi alphabet. For a phoneme in a cell, the corresponding IPA symbol and the ASCII representation used in the database are shown in second and third row of the cell respectively.**

The first 4 consonants in each row of table 2. Belong to non-nasalized category. The first two are unvoiced, the next two are voiced and the fifth one is nasal consonants. The first and third column are of the unaspirated types whereas the second and fourth are of aspirated. With rows-wise and column-wise arrangement of the consonants as in table 2. the pattern of classifications according to the place of articulation and manner of production. Each cell in the Table 2. represent a phoneme and has 3 rows; the first row is the Devnagri script, the second corresponding symbol of International Phonetic Alphabet (IPA) ,and the row the roman script used to label the phoneme in a spoken Hindi sentence. We have used English letters to represent each of the consonant .For instance the letter 'H' is used to denote the sounds 'th',and 'W' is used to denote its voice version 'dh'. Some examples of the unaspirated consonants with other vowel-endings are given in Table 3.

| Utterance Consonants-vowel | Pronounced as in |
|---|---|
| KA | CUT |
| BE | be |
| GA | GUM |
| HE | THIN |

**Table 3.Example of some monosyllabic utterances**

The four semi-vowels are [j],[r],[l],[w] and four fricatives are [s],[S],[s],[h],;last row of the table 2.These are characterized by different places of articulation, in addition to other features such as roll,frills,etc.

The effect of phonetic context on the spectra of phonemes is illustrated in Figure 4.This Figure shows the time waveform and spectrogram of the sentence *"HUM SAB EK HAI"*.Spectrogram is a method of displaying spectrum of a signal as a function of time. Here time and frequency are plotted along x and y axis respectively. The amplitude at a given frequency and at a given time is indicated by the darkness of the corresponding pixel.

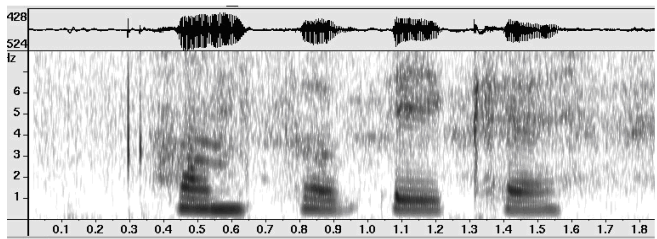

**Figure 4.The time waveform and spectrogram of the sentence "HUM SAB EK HAI".**

Dark areas of spectrogram show high intensity. Voiced segments are much louder than unvoiced. Horizontal dark bands are the formants peaks. The word **HAI** consists of *AI* which is a diphthong, combination of two vowels.

## 5. Conclusion

There is an urgent requirement for developing human oriented interface to computer. Spoken languages are still the means of communication used by humans. The ability to interact with computer in one's native language is very relevant to a multi-lingual country such as India. This paper gives the basic idea of speech recognition system and gives you the basic idea of Hindi acoustic –phonetics characteristic.

## 6. References

[1] *Samudravijaya K*, **Role of Linguistics in Spoken Language Technology,** Invited talk at the 29th All India Conference of Dravidian Linguistics Thiruvananthapuram, February 10-12, 2002.
[2] R. K. Aggarwal and Mayank Dave "Implementing a Speech Recognition System    Interface for Indian Languages" IJCNLP-08 Workshop on NLP for Less Privileged, IIIT, Hyderabad, 11 January 2008.
[3] N. Rajput M. Kumar and A. Verma. A large-vocabulary continuous speech recognition system for Hindi. IBM Journal for Research and Development.
[4] Internet-accessible speech recognition technology. http://www.cavs.msstate.edu/ hse/ies/projects/speech/index.html.
[5] Cmu sphinx - open source speech recognition engines.
[6] Samudravijaya K and Maria Barot. A comparison of public domain software tools for speech recognition. Pages 125–131. Workshop on Spoken Language Processing, 2003.
 [7] Lawrence Rabiner, Biing-Hwang Juang and B.Yegnanarayana,"*Fundamentals of   Speech Recognition"*, Pearson Education, 2009.
[8] Samudravijaya *K*, Speech and Speaker Recognition: A tutorial, Proc. Int. Workshop on Tech. Development in Indian Languages, Kolkata, Jan 22-24, 2003.
[9] Joseph W.Picone.Sep. 1993. Signal Modeling Technique in Speech Recognition.IEEE Proc...
[10] M.Karanjanadecha and Stephen A.Zahorian.1999, Signal Modeling for Isolated Word Recognition, ICASS SP Vol 1, p 293-296.
[11] Sandipan Chakroborthy and Goutam Saha, "Improved Text-Independent Speaker Identification Using Fused MFCC & IMFCC Feature Sets Based on Gaussian Filter," International Journal of Signal Processing, vol.5, no.1, pp.11-19, 2009.
[12] R. K. Aggarwal and Mayank Dave "Implementing a Speech Recognition System    Interface for Indian Languages" IJCNLP-08 Workshop on NLP for Less Privileged, IIIT, Hyderabad, 11 January 2008.
[13] L.Rabiner and B.H. Juang, "Fundamental of Speech Recognition,"PTRPrentice Hall, Englewood Cliffs, New Jersey, 1993.
[14] Murty &Yegna, "Combining Evidence from Residual Phase and MFCC features for Speaker Recognition, IEEE Signal Processing Letters, V13-1, (2006).